\documentclass[a4paper,twoside]{article}

\usepackage{epsfig}
\usepackage{calc}
\usepackage{amssymb}
\usepackage{amstext}
\usepackage{amsmath}
\usepackage{amsthm}
\usepackage{multicol}
\usepackage{pslatex}
\usepackage{apalike}
\usepackage{subcaption}
\usepackage[noend]{algpseudocode}
\usepackage{caption}
\usepackage{algorithm}
\usepackage{graphicx}
\usepackage{csquotes}
\usepackage{SCITEPRESS}     
\usepackage{url}
\usepackage{flushend}
\usepackage{coverpage}

\begin{document}

\title{Improved Discrete RRT for Coordinated Multi-Robot Planning}

\author{\authorname{Jakub Hv\v{e}zda\sup{1,2}, Miroslav Kulich\sup{1} and Libor P\v{r}eu\v{c}il\sup{1}}
\affiliation{\sup{1}Czech Institute of Informatics, Robotics, and Cybernetics,
Czech Technical University in Prague, Prague, Czech Republic}
\affiliation{\sup{2}Department of Cybernetics, Faculty of Electrical Engineering, Czech Technical University in Prague, Czech Republic}
\email{hvezdjak@fel.cvut.cz, kulich@cvut.cz, preucil@cvut.cz}
}


\abstract{
This paper addresses the problem of coordination of a fleet of mobile robots -- the problem of finding an optimal set of collision-free trajectories for individual robots in the fleet.
Many approaches have been introduced during last decades, but a minority of them is practically applicable, i.e. fast, producing near-optimal solutions, and complete.
We propose a novel probabilistic approach based on the Rapidly Exploring Random Tree algorithm (RRT) by significantly improving its multi-robot variant for discrete environments. 
The presented experimental results show that the proposed approach is fast enough to solve problems with tens of robots in seconds. 
Although the solutions generated by the approach are slightly worse than one of the best state-of-the-art algorithms presented in~\cite{terMors2010}, it solves problems where ter Mors's algorithm fails. 
}

\mauthor{Jakub Hv\v{e}zda, Miroslav Kulich, and Libor P\v{r}eu\v{c}il}
\published{{\it Proceedings of the 15th International Conference on Informatics in Control, Automation and Robotics} - Volume 2: ICINCO 2018, 
ISBN 978-989-758-321-6, pages 171-179.}
\DOI{10.5220/0006865901810189}
\original{http://www.scitepress.org/DigitalLibrary/Link.aspx?doi=10.5220/0006865901810189}
\coverpage

\onecolumn \maketitle \normalsize \vfill

\section{\uppercase{Introduction}}
Recent advances in mobile robotics and increased deployment of robotic systems in many practical applications lead to intensive research of multi-robot systems. 
One of the most important problems is the coordination of trajectories of individual robots/agents in such systems: given starting and destination positions of the robots, we are interested in finding their trajectories that do not collide with each other, and the overall cost is minimized. 
An optimization criterion can be a sum of lengths of the individual trajectories or the time the last robot reaches its destination position.

\begin{figure}[t]
    \centering
    \includegraphics[width=\columnwidth]{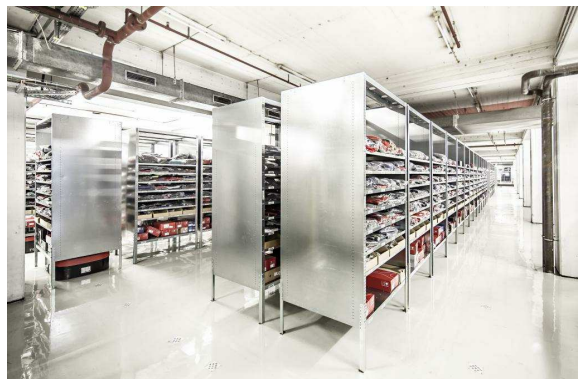}
      \caption{Automated warehouse: G-COM system by Grenzebach (\url{https://www.grenzebach.com}) in a costumer application.}
    \label{fig:gcom}
\end{figure}

Several fields of the industry such as airports are nowadays faced with a higher increase in traffic than the actual capacity. 
This leads to reliance on path optimizations to increase their throughput.
Another typical application where multi-robot coordination plays an important role might be planning in an automated warehouse, see Fig.~\ref{fig:gcom}, where autonomous robots effectively deliver desired goods from/to given positions.
 
Multi-robot path planning and motion coordination has been studied since the 1980s, and many techniques have been developed during this period, see~\cite{Parker2009} for a nice overview. This problem (formulated as the warehouseman's problem) was proved to be PSPACE-complete~\cite{Hopcroft1984}. For the case where robots move on a predefined graph complexity of the problem can be reduced, nevertheless, it is still NP-hard~\cite{Goldreich11}, which means that optimal solutions cannot generally be found in a reasonable time for non-trivial instances (e.g., for a number of robots in order of tens).

Solutions to the problem consider either coupled or decoupled approaches. Centralized (coupled) approaches consider the multi-robot team as a multi-body robot for which classical single-robot path planning can be applied in composite configuration space. Traditional centralized methods are based on complete (i.e., the algorithm finds a solution if it exists or reports that no solution exists otherwise) and optimal classical algorithms and provide optimal solutions~\cite{Latombe1991},~\cite{Lavalle1998},~\cite{Ryan2008}. However, these approaches require computational time exponential in the dimension of the composite configuration space, so they are appropriate for small-sized problems only. This drawback leads to the development of methods that prune the search space. For instance, Berg et al.~\cite{Berg2009} decompose any instance of the problem into a sequence of sub-problems where each subproblem can be solved independently from the others. The Biased Cost Pathfinding~\cite{Geramifard2006biased} employs generalized central decision maker that resolves collision points on paths that were pre-computed independently per unit, by replanning colliding units around the highest priority unit. Another approach is to design an algorithm based on a specific topology describing the environment. \cite{Peasgood08} present a multi-phase approach with linear time complexity based on searching a minimum spanning tree of the graph, while an approach for grid-like environments is introduced in~\cite{Wang2008}. A flow-annotated search graph inspired by two-way roads is built to avoid head-to-head collisions and to reduce the branching factor in search. Nevertheless, the computational complexity is still high (e.g.,~\cite{Berg2009} solves a problem with 40 robots in 12 minutes,~\cite{Wang2008} needs approx. 30 seconds for 400 robots).

On the contrary, decoupled methods present a coordination phase separated from the path planning phase. These approaches provide solutions typically in orders of magnitude faster times than coupled planners, but these solutions are sub-optimal. Moreover, the decoupled methods are often not complete as they may suffer from deadlocks. These approaches are divided into two categories: path coordination techniques and prioritized planning. Path coordination considers tuning the velocities of robots along the precomputed trajectories to avoid collisions~\cite{LaValle1998_OMP},~\cite{Simeon2002}, while prioritized planning computes trajectories sequentially for the particular robots based on the robots' priorities. Robots with already determined trajectories are considered as moving obstacles to be avoided by robots with lower priorities~\cite{VandenBerg2005},~\cite{Bennewitz2001},~\cite{Cap2015}. 
A similar idea was presented in \cite{terMors2010}, where adaptation of the A* algorithm is sequentially planning on a graph augmented by information in which time intervals the particular nodes are occupied by already processed robots.

Several computationally efficient heuristics have been introduced recently enabling to solve problems for tens of robots in seconds. Windowed Hierarchical Cooperative A* algorithm (WHCA*) employs heuristic search in a space-time domain based on hierarchical A* limited to a fixed depth~\cite{Silver2005}. Chiew~\cite{Chiew2010} proposes an algorithm for $n^2$ vehicles on a $n\times n$ mesh topology of path network allowing simultaneous movement of vehicles in a corridor in opposite directions with computational complexity $O(n^2)$. Luna and Bekros~\cite{Luna2011} present a complete heuristics for general problems with at most $n-2$ robots in a graph with n vertices based on the combination of two primitives - \enquote{push} forces robots towards a specific path, while \enquote{swap} switches positions of two robots if they are to be colliding. An extension which divides the graph into subgraphs within which it is possible for agents to reach any position of the subgraph, and then uses \enquote{push}, \enquote{swap}, and \enquote{rotate} operations is presented in~\cite{DeWilde2014}.
Finally, Wang and Wooi~\cite{Wang2015} formulate multi-robot path planning as an optimization problem and approximate the objective function by adopting a maximum entropy function, which is minimized by a probabilistic iterative algorithm.

Although many of the approaches mentioned above have nice theoretical properties, the most practically usable algorithm is probably Context-Aware Route Planning (CARP) presented in~\cite{terMors2010} as it is fast, produces solutions of high quality, and although it is not complete, it finds a solution for a large number of practical setups.  

\cite{drrt} presented MRdRRT algorithm, which is a probabilistic approach for path planning on predefined structures for relatively small number of robots inspired by RRT algorithm \cite{Lavalle1998}. 
\cite{Dobson2017} improves upon MRdRRT by presenting its optimal variant.

In this paper, we present a probabilistic approach which extends and improves a discrete version of Rapidly-Exploring Random Tree (RRT) for multiple robots~\cite{drrt}.
Our approach focuses mainly on scalability with increasing number of agents as well as improving the quality of solution compared to \cite{Dobson2017} that presents the optimal version of the dRRT algorithm but keeps the number of robots relatively low.
We show that the proposed extensions allow solving problems with tens of robots in times comparable to CARP with a slightly worse quality of results. 
On the other hand, the proposed algorithm finds solutions also for setups where CARP fails.

The rest of the paper is organized as follows. 
The multi-agent path-finding problem is presented as well as the used terms are defined in Section~\ref{sec:problem}.
The multi-robot discrete RRT algorithm and the proposed improvements are described in Section~\ref{sec:algorithm}, while performed experiments, their evaluation, and discussion are presented in Section~\ref{sec:experiments}.
Finally, Section~\ref{sec:conclusion} is dedicated to concluding remarks.

\section{\uppercase{Problem definition}}
\label{sec:problem}
Multi-agent pathfinding/coordination is a problem that is concerned about finding paths for multiple agents from their given start locations to their target locations without colliding with each other or obstacles in the environment while also optimizing a global cost function.

To specify the problem more precisely, assume:

\begin{itemize}
\item A set of \textit{k} homogenous agents each labeled $a_1, a_2,..., a_k$.    
\item A graph $G(V,E)$ where $|V| = N$. The vertices $V$ of the graph are all possible agent's locations, while $E$ represents a set of all possible agent's transitions between the locations.
\item A start location $s_i \in V$ and a target location $t_i \in V$ of each agent.
\end{itemize}

The aim is to find a set of collision-free trajectories on $G(V,E)$ each of them specifying locations of an individual agent at all time moments so that agents are at their start locations initially and at their goal locations finally.
Note that the time is discretized into time moments to simplify the problem. 

The following paragraphs explain key used terms and specify additional constraints to the generated trajectories.

\subsection{Actions}
Every agent can perform two types of action at each time point:
It can either move into one of the neighboring nodes, or it can wait at its current location.
Every algorithm can make different assumptions regarding the cost of these actions, but we assume that staying idle has zero cost of distance traveled, but costs time.
Furthermore, once an agent reaches its target location, it waits at this location for other agents to finish.

\subsection{Constraints}
The main constraints on agent movement assumed in this paper are:

\begin{itemize}
\item No two agents $a_1$ and $a_2$ can occupy the same vertex $v \in V$ at the same time.
\item Assume two agents $a_1$, $a_2$ located in two neighboring nodes $v_1,v_2 \in V$ respectively, they can not travel along the same edge $(v_1,v_2)$ at the same time in opposite directions.
In other words, two neighboring agents cannot swap positions.
However, it is possible for agents to follow one another assuming that they do not share the same vertex or edge concurrently.
For example, if the agent $a_1$ moves from $v_2 \in V$ to $v_3 \in V$ then the agent $a_2$ can move from $v_1 \in V$ to $v_2 \in V$ at the same time.
\end{itemize}

\subsection{Composite configuration space}
The \textit{composite configuration space} $\mathcal{G} = (\mathcal{V},\mathcal{E})$ is a graph that is defined as follows.
The vertices $\mathcal{V}$ are all combinations of collision-free placements of \textit{m} agents on the original graph $G$.
These vertices can also be viewed as $m$ agent configurations $C = (v_1,v_2,...,v_m)$, where an agent $a_i$ is located at a vertex $v_i \in G$ and the agents do not collide with each other.
The edges of $\mathcal{G}$ can be created using either Cartesian product or Tensor product.
For the purposes of this paper the Tensor product is used because is allows simultaneous movement movement of multiple agents and thus for two $m$ agent configurations $C = (v_1,v_2,...,v_m)$, $C' = (v'_1,v'_2,...,v'_m)$ the edge $(C,C')$ exists if $(v_i,v'_i) \in E_i$ for every $i$ and no two agents collide with each other during the traversal of their respective edges.

The distance between two neighboring nodes $C_1=(v_{11},v_{12}, ... , v_{1n})$ and $C_2=(v_{21},v_{22}, ... , v_{2n})$ in a composite roadmap is calculated as the sum of Euclidean distances $d$ between the corresponding nodes:

$$\delta\left(C_1,C_2\right) = \sum_{i = 0}^{n} d(v_{1i},v_{2i})$$

\section{\uppercase{Proposed algorithm}}
\label{sec:algorithm}
\subsection{Discrete RRT} \label{sec:drrt}
A discrete multi-robot rapidly-exploring random tree (MRdRRT)\cite{drrt} is a modification of the RRT algorithm for pathfinding in an implicitly given graphs embedded in a high-dimensional Euclidean space.

 

Just like RRT, the MRdRRT grows a tree $\mathcal{T}$ rooted in the vertex $s$ representing start positions of the robots in a composite configuration space $\mathbb{R}^d$ by iteratively adding new points to the tree while also trying to connect to the goal configuration $t$ without violating any constraints, e.g. collision with the environment.
The growth is achieved by randomly sampling a point $u$ in the composite configuration space and then extending the tree towards this sample.
Note that vertices newly added to the tree are taken from $G$: given a sample $u$ and the node $v\in V$ nearest to it, the best neighbor $v'\in V$ has to be found. 
To generate neighbor nodes of already visited nodes MRdRRT uses a technique called oracle. Without loss of generality consider that $G$ is embedded in $\left[0,1\right]^d$.
For two points $v,v' \in \left[0,1\right]^d$ the $\rho\left(v,v'\right)$ denotes a ray that begins in $v$ and goes through $v'$.
$\angle_v\left(v',v''\right)$ given three points $v,v',v'' \in \left[0,1\right]^d$ denotes the (smaller) angle between $\rho\left(v,v'\right)$ and $\rho\left(v,v''\right)$.
The way the oracle is used is given sample point $u$ it returns the neighbor $v'$ of $v$ such that the angle between rays $\rho\left(u,v'\right)$ and $\rho\left(v,v'\right)$ is minimized. This can be defined as

$$\mathcal{O}_D\left(v,u\right) := \underset{v' \in V}{\mathrm{argmin}} \left\lbrace \angle_v\left(u,v'\right)|\left(v,v'\right) \in E \right\rbrace.$$

It is possible that the tree will if given sufficient time, eventually reach $t$ during the expansion phase. 
However it is unlikely for larger problems.
MRdRRT, therefore, attempts to connect the newly added node with $t$ employing so-called {\em local connector} which is successful for restricted problems only, but fast, so it can be run often.

\subsection{Proposed improvements}
Although the original MRdRRT can solve path-finding problems for several robots, the realization of its particular steps is inefficient, which disqualifies it to deal with complex scenarios containing tens of robots.
The authors of MRdRRT present experimental results with up to 10 robots and mention that their algorithm is unable to solve problems with a significantly larger number of robots.
We, therefore, introduce several improvements to the original version. 

The original expansion phase generates random samples from a bounding box of $G$ which is inefficient in maps with tight spaces as it would not allow robots to stand still as their next action and would not find a solution in situations where standing still was required for one of the robots.
Moreover, the majority of generated points is far from a solution leading to a relatively huge growth of the tree over the configuration space and thus to the high computational complexity of the algorithm.
Instead of generating a point from $\mathbb{R}^d$, we find shortest paths for every robot separately in the preprocessing phase and after
that we compose a sample only from points $q$ for which $dist(s_i, q) + dist(q, t_i) \leq dist(s_i, t_i) + \Delta$,  where $dist$ is a distance of two points, $s_i$ and $t_i$ are start and goal positions of $i$-th robot, and $\Delta>0$ is a defined constant threshold.

The original oracle generates a sample and checks it for collisions, which is inefficient as many samples are discarded.
Our version iterates over positions of all robots $v_i$ and tries to generate a new step $v_i'$  for each of them towards the sample point $u_i$ while avoiding collisions and also minimizing the angle $(u_i,v_i,v_i')$ by keeping a list of collision configurations that need to be avoided.

Another proposed improvement is the use of the CARP algorithm~\cite{terMors2010} as a local connector as well as a random shuffling of the order in which CARP attempts to plan trajectories of individual agents to their desired locations. 
This algorithm creates a free time window graph on which agents find the shortest paths one by one while updating the free time window graph with their paths so that collisions are avoided.

\begin{algorithm}
\caption{Improved MRdRRT algorithm}\label{alg:rrt*_overview}
\begin{algorithmic}[1]
\State {$\mathcal{T}.init\left(s\right)$}
\Loop
    \State {$EXPAND\left(\mathcal{T}\right)$}
    \State {$REWIRE\left(\mathcal{T},v'\right)$}
    \State {$\mathcal{P}\leftarrow CONNECT\_TO\_TARGET\left(\mathcal{T},t\right)$}
    \If {$not\_empty(\mathcal{P})$}
        \Return $RETRIEVE\_PATH\left(\mathcal{T},\mathcal{P}\right)$
    \EndIf
\EndLoop
\end{algorithmic}
\end{algorithm}

The last set of modifications to the algorithm is the addition of steps inspired by RRT* algorithm \cite{rrt*} which include the new rewiring step and modification of expansion step, see Alg.~\ref{alg:rrt*_overview}.
At the start of the algorithm, the tree is initialized with the node that contains the initial configuration of agents (line 1).
The main loop of the algorithm then starts with the newly modified expansion phase (line 3).
After a new node is added the rewiring step is called (line 4) that attempts to revise the structure of the tree to improve path length to the root.
The algorithm finally tries to connect the newly added node with the goal configuration.
If it succeeds the algorithm returns the found path.

The change to the expansion phase, Alg.~\ref{alg:mrmrrt*_expand}, consists of connecting the new node $v'$ to a node already in the tree $\mathcal{T}$ that minimizes the distance traveled from the initial configuration $s$.
The additional step of the expansion phase in the original RRT* consists of checking nodes in the radius around the new node $v'$ for the best predecessor and then connecting $v'$ to it.
However, in the multi-agent discrete scenario (Alg. \ref{alg:mrmrrt*_expand}) the computational requirements to perform a similar task are much higher because it would require to run a local connector method on each node in the radius and then perform the distance to root check.
The expansion phase was thus modified so that it employs nearest neighbor search instead of radius (line 2).
The key difference is that in the first step of expansion the random sample $u$ (line 1) is generated, but after that the new node $v'$ is not created from the nearest neighbor of $u$.
Instead, $N$ nearest neighbors of $u$ are iterated over (lines 6-11) and a new node $v'$ is generated from each of them using the oracle $\mathcal{O}_D$, but not added into the tree.
Each $v'$ is checked for the distance traveled through the tree $\mathcal{T}$ towards the root $s$ and only a node that minimizes this distance is connected to its corresponding predecessor.

\begin{algorithm}
\caption{Improved MRdRRT EXPAND$\left(\mathcal{T},r\right)$}\label{alg:mrmrrt*_expand}
\begin{algorithmic}[1]
    \State {$u \leftarrow RANDOM\_SAMPLE\left(\right)$}
    \State {$NNs \leftarrow getNearestNeighbours(u)$}    
    \State {$v'_{pred} = -1$}
    \State {$d_{best} = \infty$}
    \State {$v'_{best} = \emptyset$}
    \For {$c \in NNs$}
        \State {$v' \leftarrow \mathcal{O}_D\left(c,u\right)$}
        \If {$l_{\mathcal{T}}\left(c\right) + \delta\left(c,v'\right) < d_{best}$}
            \State {$d_{best} = l_{\mathcal{T}}\left(c\right) + \delta\left(c,v'\right)$}
            \State {$v'_{pred} = c$}
            \State {$v'_{best} = v'$}
        \EndIf
    \EndFor
    \State {$\mathcal{T}.add\_vertex\left(v'_{best}\right)$}
    \State {$\mathcal{T}.add\_edge\left(v'_{pred},v'_{best}\right)$}
\end{algorithmic}
\end{algorithm}

\begin{figure}[htb]
    \begin{subfigure}[b]{\columnwidth}
            \centering
            \includegraphics[width=0.95\columnwidth]{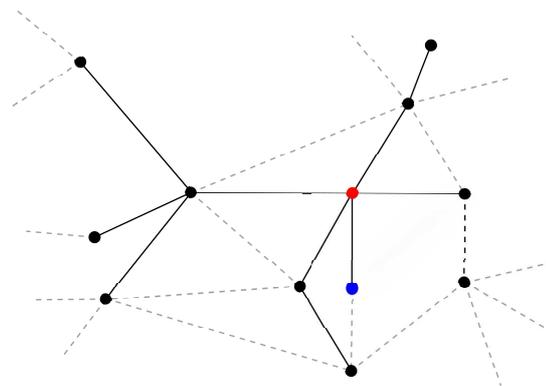}
            \caption{Initial state of the tree. The root node is coloured red, while the newly added node $v'$ is coloured blue.}
            \label{fig:rewire_1}
    \end{subfigure}  

    \bigskip    

    \begin{subfigure}[b]{\columnwidth}
            \centering
            \includegraphics[width=0.95\columnwidth]{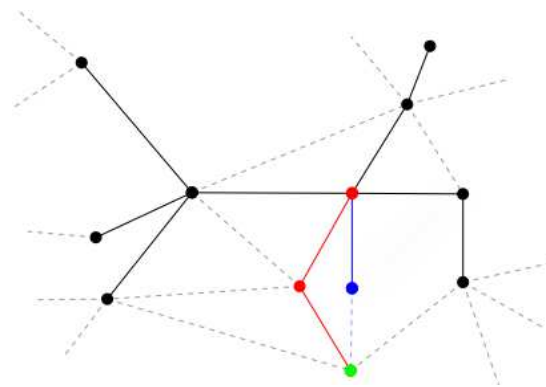}
            \caption{The path through the newly added node $v'$(blue) to one of its nearest neighbors (green) is shorter than the current path to this node (red).}
            \label{fig:rewire_2}
    \end{subfigure}

    \bigskip    
    
    \begin{subfigure}[b]{\columnwidth}
            \centering
            \includegraphics[width=0.95\columnwidth]{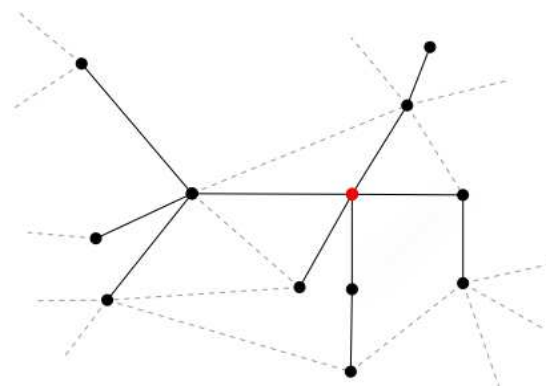}
            \caption{The tree is revised.}
            \label{fig:rewire_3}
    \end{subfigure}
    \caption{Example of the rewiring procedure.} 
    \label{fig:rewire}
\end{figure}

The rewiring step of RRT* locally revises a structure of $\mathcal{T}$ by checking whether nodes within the radius $r$ around a newly added node $v'$ have the distance traveled towards the root node shorter when they had $v'$ as their predecessor.
This step was modified for the use in a multi-agent discrete case (Alg.~\ref{alg:mrdrrt*_rewire}) by omitting the radius and iterating over $N$ nearest neighbors of $v'$ instead.
Because these neighboring configurations $c$ might not be direct neighbors of $v'$ in the composite graph $G$, the local connector is used to obtain a path between these two nodes (line 3).
If the local connector fails to find the path, the neighbor is immediately skipped (lines 4-5).
In the case the local connectors successes in finding a path $p$ between $v'$ and $c$ it is checked whether the length of the path from the root to $v'$ concatenated with the path $p$ and the node $c$ is shorter than the distance traveled through $\mathcal{T}$ from the root to $c$(lines 5-7).
If it is shorter, then all nodes of $p$ are added to $\mathcal{T}$.
The first node of $p$ is connected as the successor of $v'$ and the last node of $p$ is chosen as a new predecessor of $c$.
An example of the rewiring step is displayed in Fig.~\ref{fig:rewire}. 

\begin{algorithm}
\caption{REWIRE$\left(\mathcal{T},v'\right)$} \label{alg:mrdrrt*_rewire}
    \begin{algorithmic}[1]
        \State {$NNs \leftarrow getNearestNeighbours(v')$}    
        \For {$c \in NNs$}
            \State {$p \leftarrow LOCAL\_CONNECTOR\left(v',c\right)$}
            \If {$p \leftarrow \emptyset$}
                \State {$Continue$}
            \EndIf
            \State {$n \leftarrow LastNode\left(p\right)$}
            \If {$l_{\mathcal{T}}\left(v'\right)+l\left(p\right)+\delta\left(n,c\right) < l_{\mathcal{T}}\left(c\right)$}            
                \State {$\mathcal{T}.add(p)$}                
                \State {$c.predecessor = n$}
            \EndIf
        \EndFor
    \end{algorithmic}
\end{algorithm}

\section{\uppercase{Experiments}}
\label{sec:experiments}

\begin{figure*}

    \begin{subfigure}[t]{0.45\linewidth}
    	\subcaption{Number of failed assignments}
        \includegraphics[width=\linewidth]{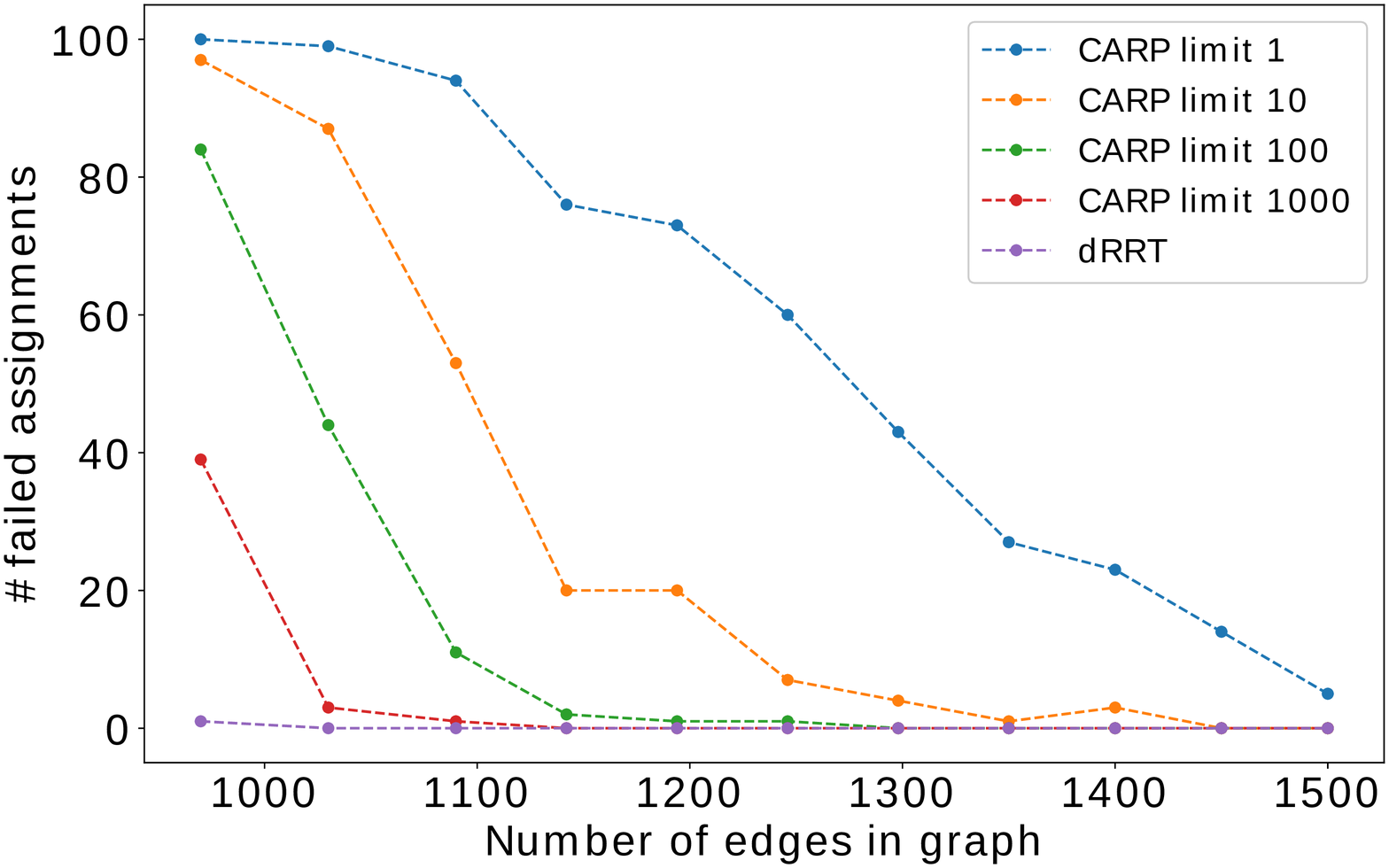}    
        \label{fig:failrate}
    \end{subfigure}
    \begin{subfigure}[t]{0.45\linewidth}
    	\subcaption{Median number of plan steps}
        \includegraphics[width=\linewidth]{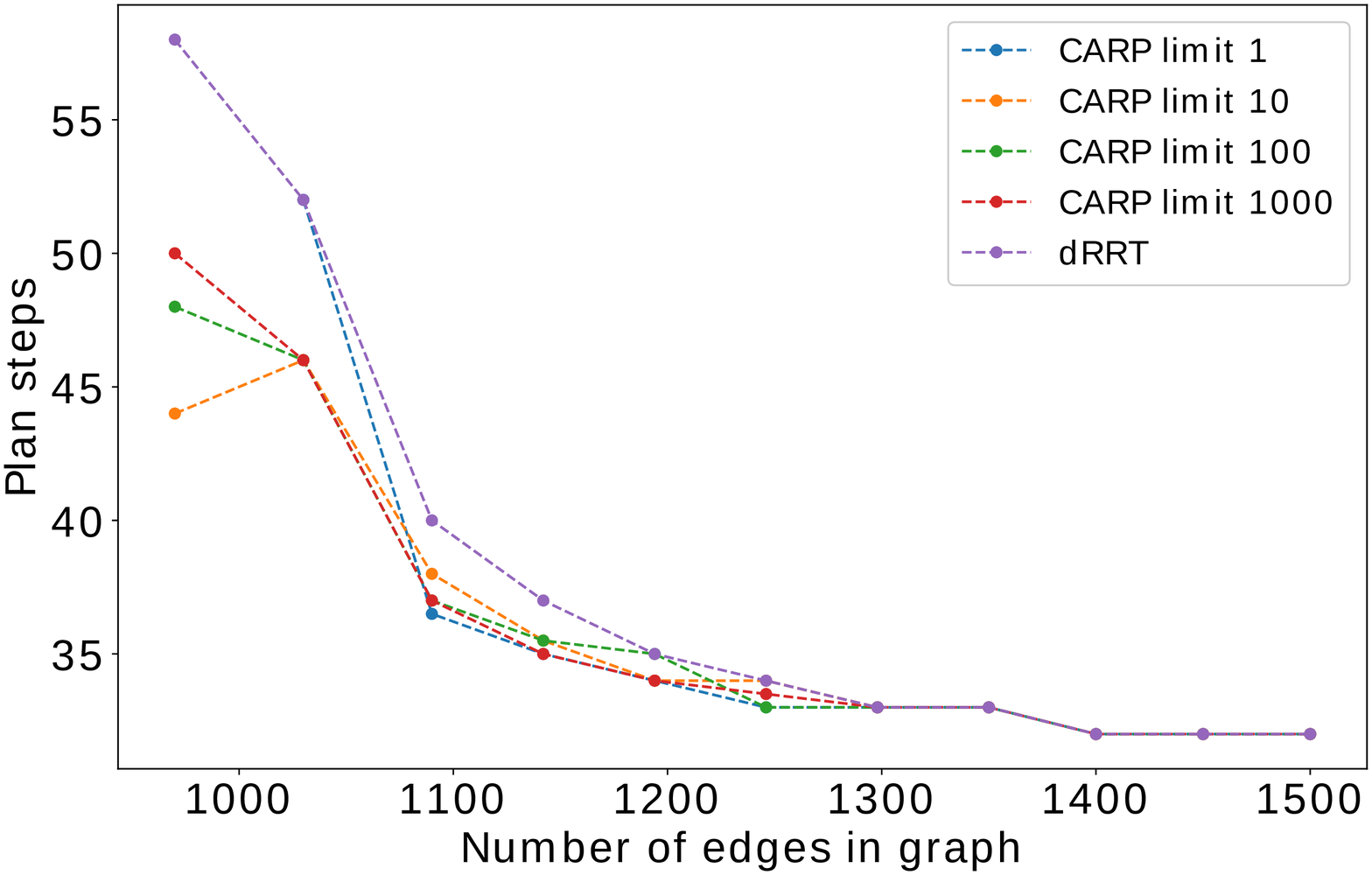}    
        \label{fig:steps}
    \end{subfigure}    
    
    \begin{subfigure}[t]{0.45\linewidth}
		\subcaption{Median iterations}        
        \includegraphics[width=\linewidth]{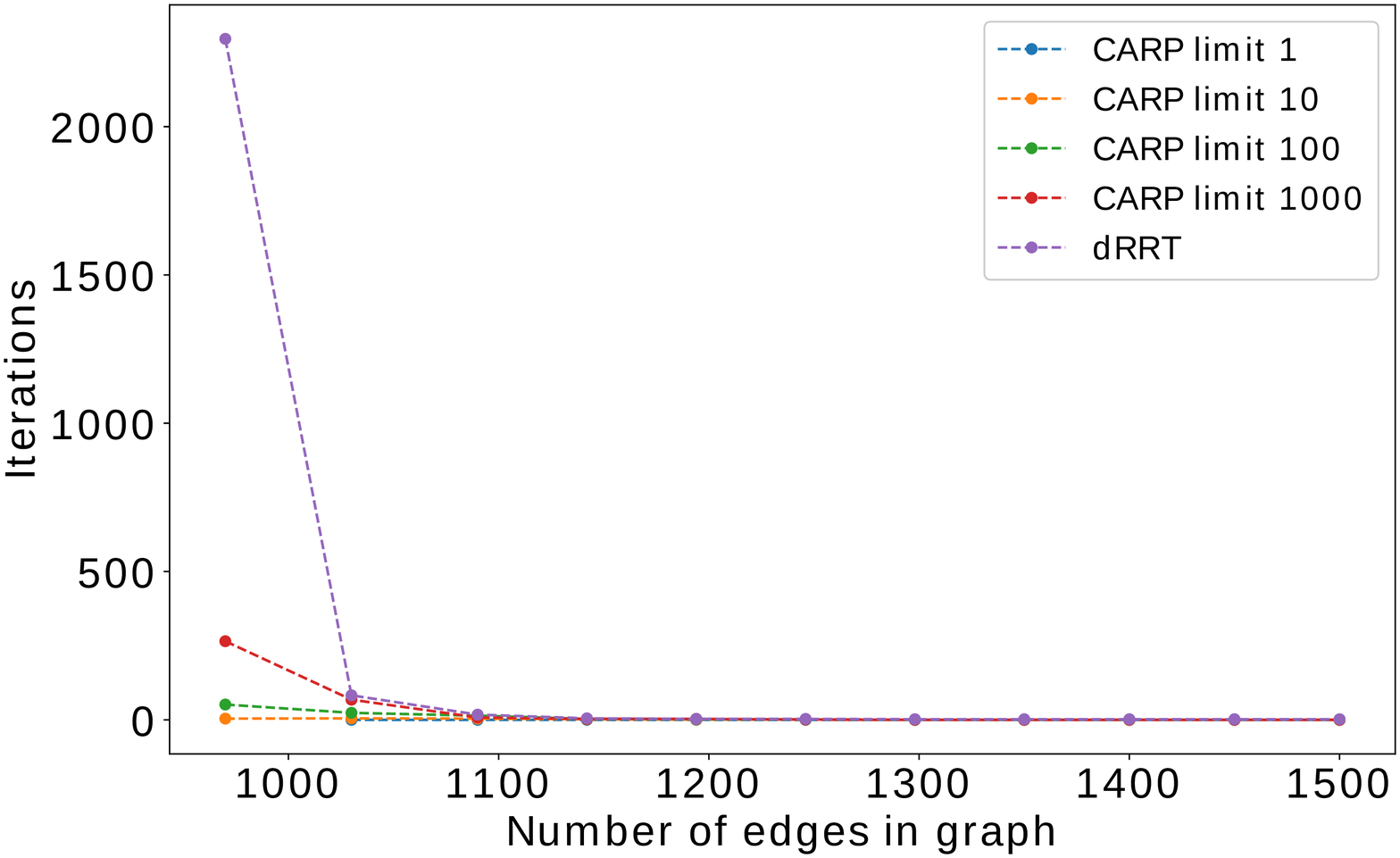}    
        \label{fig:iterations}
    \end{subfigure}    
    \begin{subfigure}[t]{0.45\linewidth}
    	\subcaption{Median time to plan}
        \includegraphics[width=\linewidth]{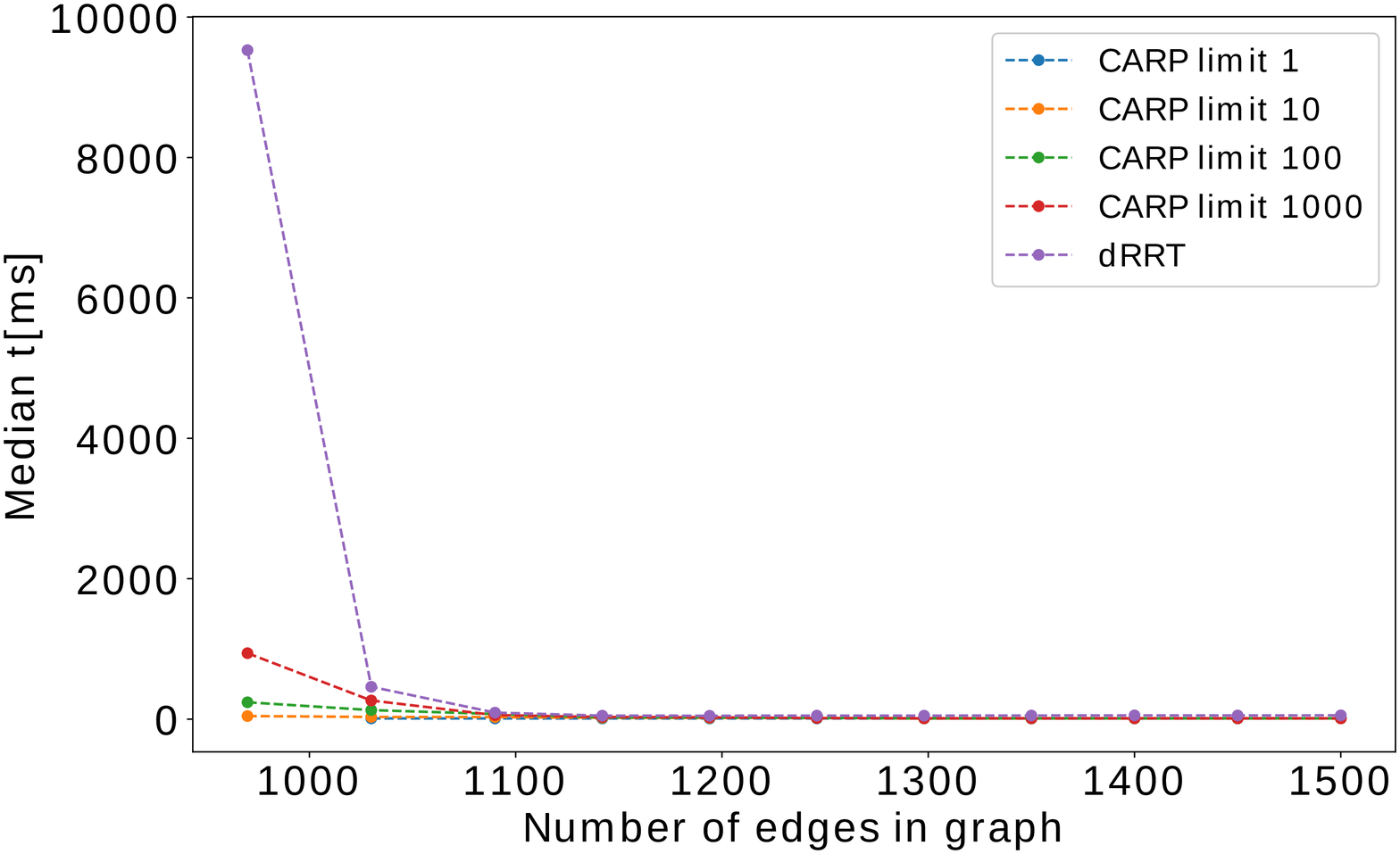}    
        \label{fig:runtime}
    \end{subfigure}
\caption{Comparison of the proposed approach (dRRT) with CARP.}
\label{fig:firstexperiment}
\end{figure*}


Performance of the proposed method has been evaluated, and comparison with the CARP algorithm~\cite{terMors2010} has been done. 
The experiments were performed on two sets of artificially created maps and assignments with the aim to compare the quality of obtained results and runtime as well as the reliability of both algorithms.

This first set of maps was created to demonstrate how both algorithms perform depending on the density of the given graph and the number of cycles in it.
The set was created by generating a random spanning tree of a $20\times 20$ grid map followed by the creation of additional maps by iteratively adding a fixed number of original edges into the spanning tree.
The experiments were thus carried out on the set consisting of 11 maps ranging from a spanning tree to the full grid. 
Furthermore, 100 different assignments for a fleet of 100 robots were created by randomly sampling start and goal nodes for each agent and each such assignment was tested on all of these maps.
The CARP algorithm was tested by giving it the limit of 1, 10, 100, 1000 attempts to find a solution where the order in which agents were planned was randomly shuffled for each attempt.

The results of this experiment can be seen in Fig.~\ref{fig:firstexperiment}. 
The first thing to notice in Fig.~\ref{fig:failrate} is that the proposed approach shows a much higher success rate even on the spanning tree, where it had its single failure.
Contrary, CARP had 100\% failure rate on the spanning tree when given only 1 attempt and 39\% when given 1000 attempts.
It can be seen in Fig.~\ref{fig:iterations} and Fig.~\ref{fig:runtime} that once the number of edges in the graph reaches 1090, the algorithms behave very similarly in terms of runtime and the needed number of iterations to find the plan.
The proposed method provides a slightly higher median of a number of steps of the resulting plan as can be seen in Fig.~\ref{fig:steps}.
This can be attributed to the fact that the median is calculated from a higher number of successful plans compared to CARP algorithm.

\begin{figure}[htb]
    \centering
    \begin{subfigure}[t]{0.4\columnwidth}
            \centering
            \includegraphics[width=\columnwidth]{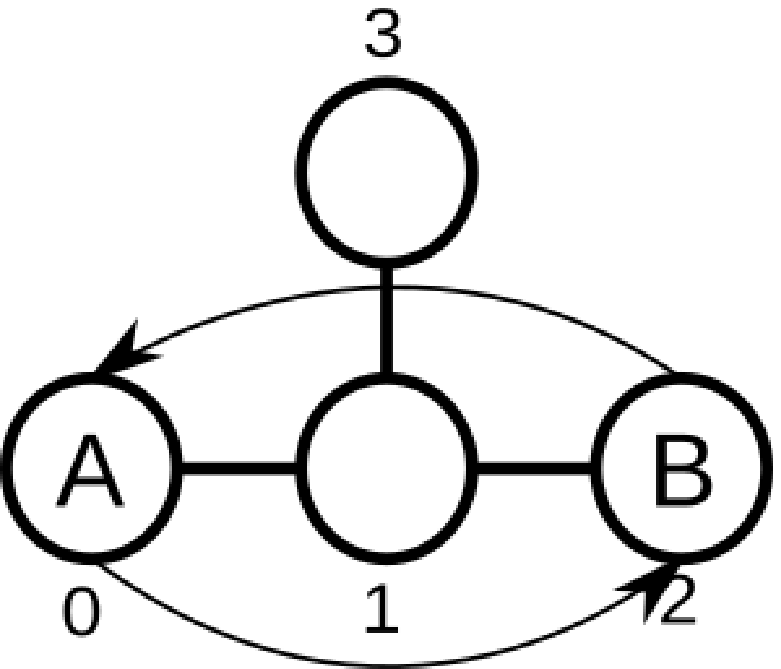}            
            \caption{Base problem.}
            \label{fig:mapgen_base}
    \end{subfigure} 
    \hfil
    \begin{subfigure}[t]{0.4\columnwidth}
            \centering
            \includegraphics[width=\columnwidth]{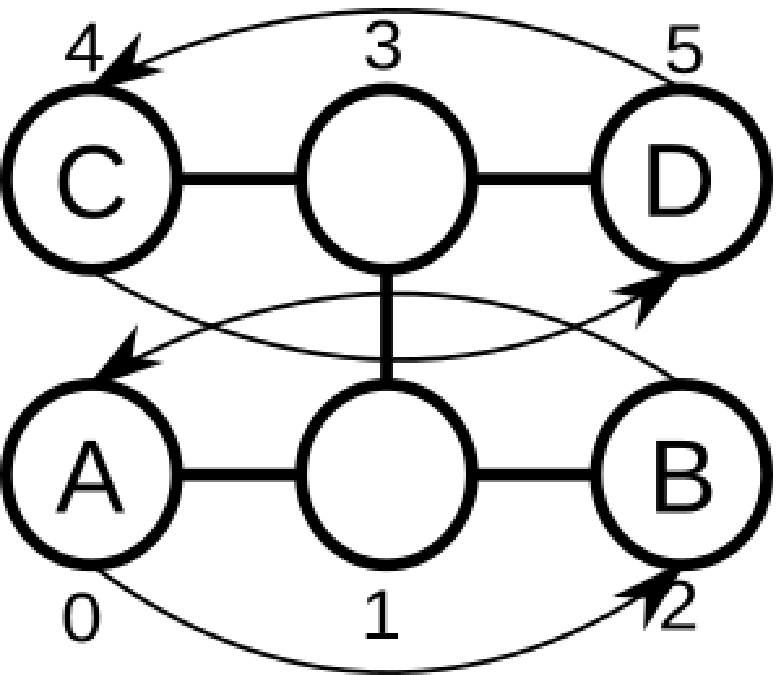}
            \caption{First type of map expansion}
            \label{fig:mapgen_1}
    \end{subfigure}
    
    \bigskip\bigskip

    \begin{subfigure}[b]{\columnwidth}
            \centering
            \includegraphics[width=0.76\columnwidth]{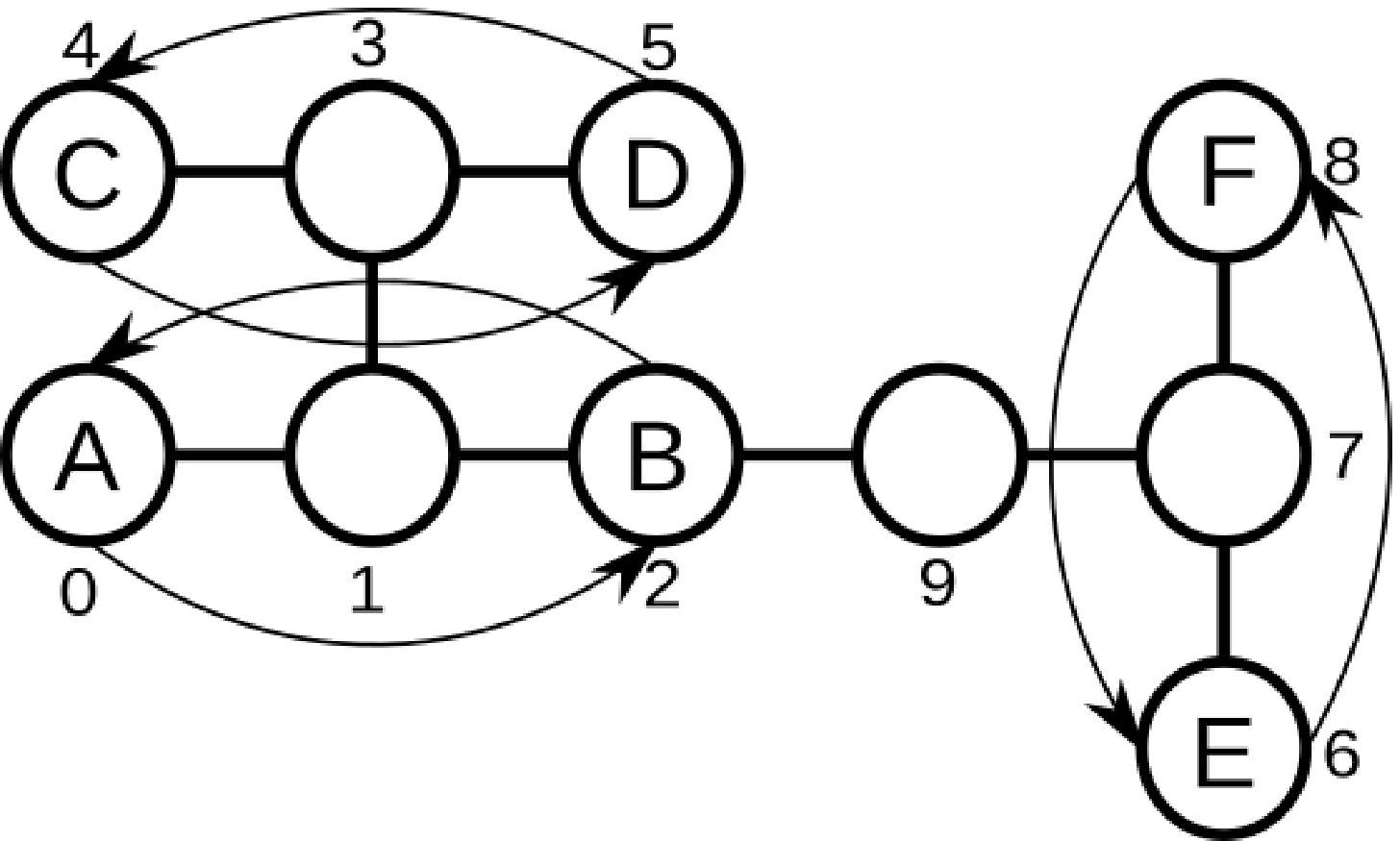}
            \caption{Second type of map expansion}
            \label{fig:mapgen_2}
    \end{subfigure}
    \caption{Map generation procedure.} 
    \label{fig:mapgen}
\end{figure}

The second set of maps was created specifically together with assignments so that the problems would be impossible to solve for the CARP algorithm. 
The maps and assignments were randomly generated by the following process:

\begin{enumerate}
\item Create a basic problem that is impossible to solve for the CARP algorithm depicted in Fig.~\ref{fig:mapgen_base}.
Arrows indicate the starting and goal positions of robots $A$ and $B$ on the graph.
CARP fails because the agents need to swap their positions while having the same distance to the only node they can use to avoid each other. Because CARP plans agents sequentially one by one while ignoring the subsequent agents, no ordering of these agents can solve this issue.
\item Pick random node that has only one edge associated with it.
\item Either add 2 nodes $A$ and $B$ to either side of this node if possible along with corresponding assignment of 2 agents -- The first agent going from $A$ to $B$ and the second one from $B$ to $A$.
The example of this step can be seen in Fig.~\ref{fig:mapgen_1}.
The alternative method to expand the map is to connect the same structure to it as in the Step 1 together with the same type of assignment, the example of which can be seen in Fig.~\ref{fig:mapgen_2}.
\item Repeat Steps 2 and 3 until the map of a required size is generated.
\end{enumerate}

\begin{figure}[htb]
    \centering
    \includegraphics[width=\columnwidth]{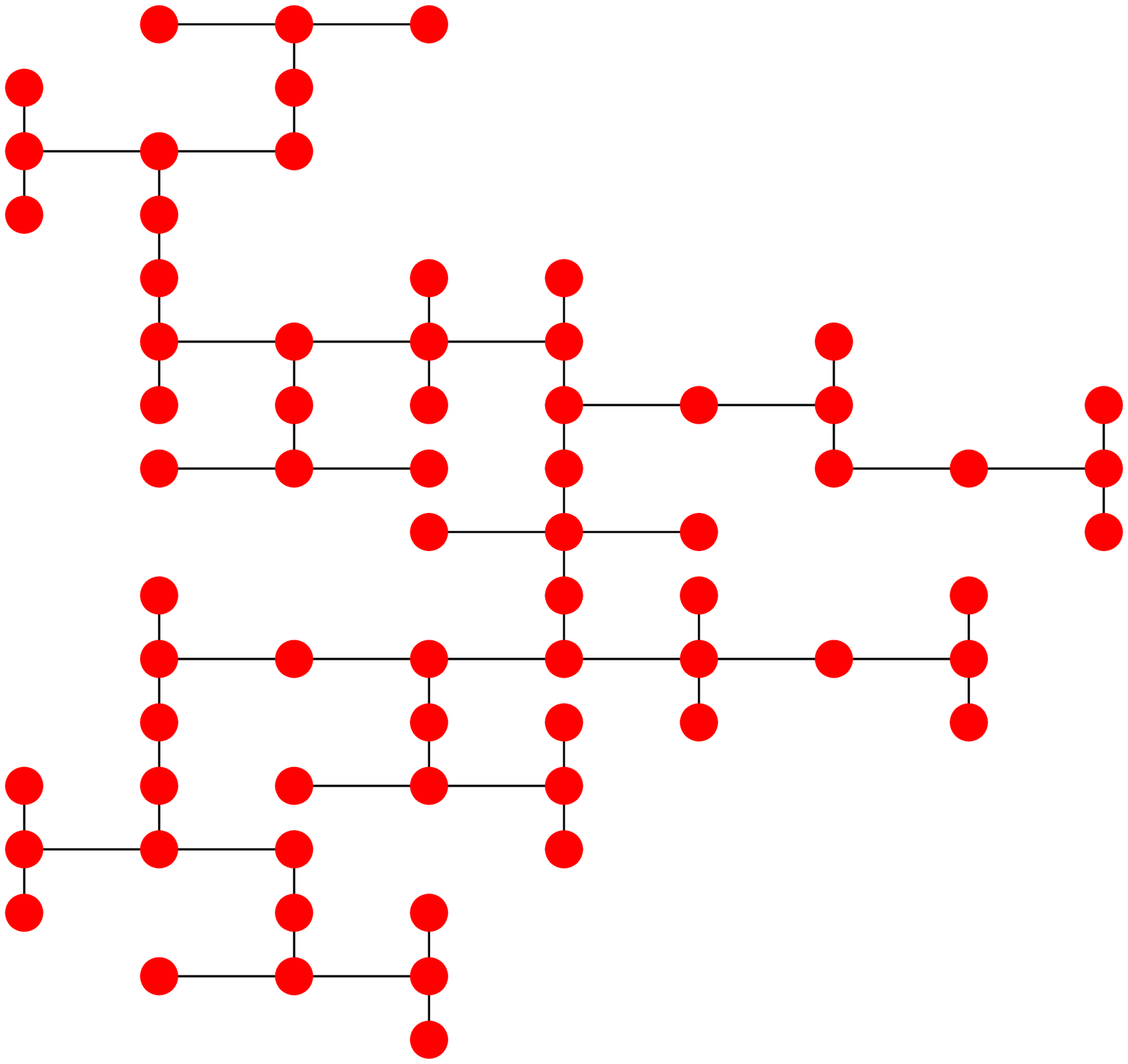}
    \caption{Example of a generated map.}
    \label{map_example}
    \end{figure}

The example of a fully generated map following the previous steps can be seen in Fig.~\ref{map_example}.

\begin{figure*}[htb]
\begin{tabular}{ccc}
\includegraphics[width=5cm,scale=1]{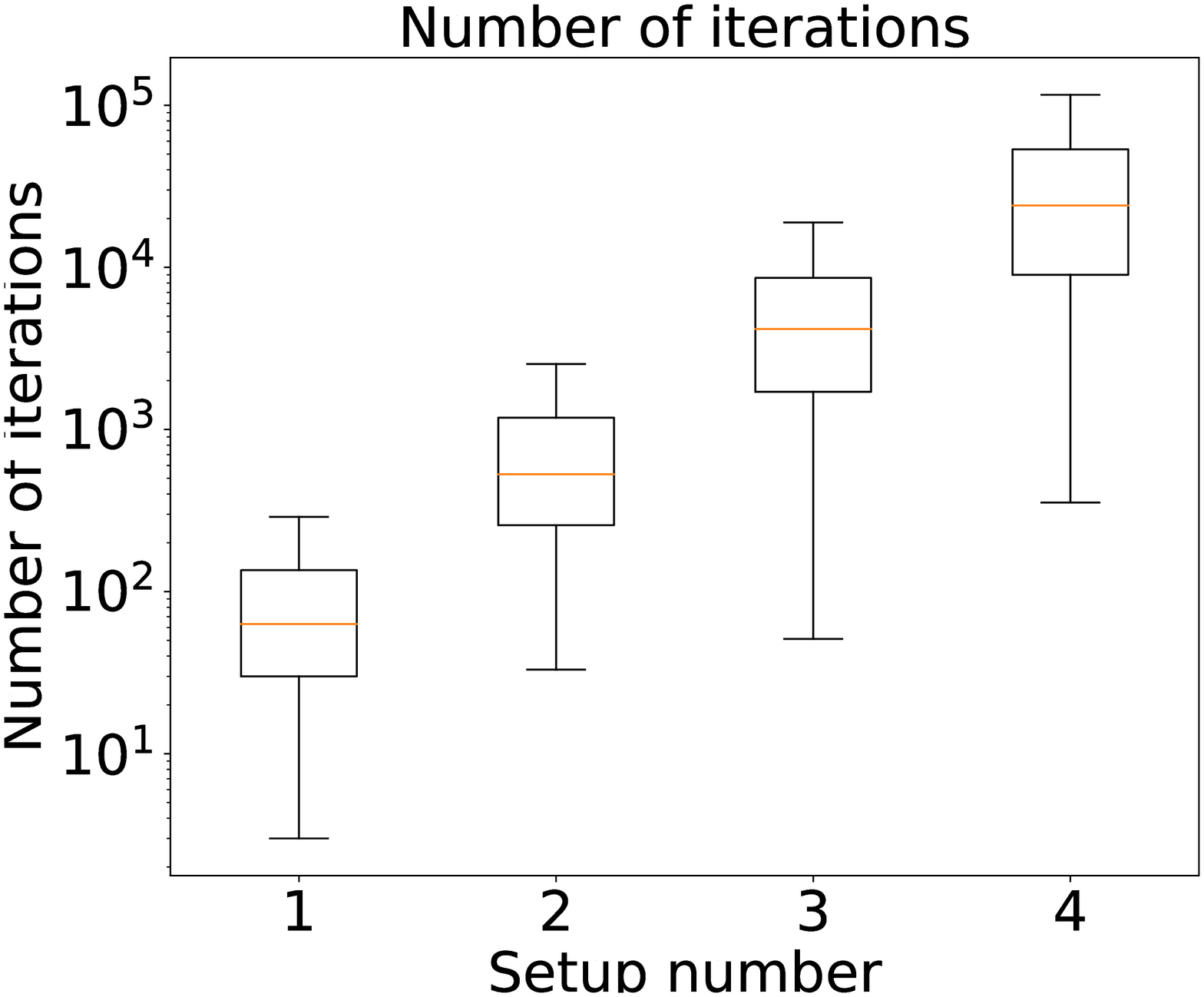} & \includegraphics[width=5cm,scale=1]{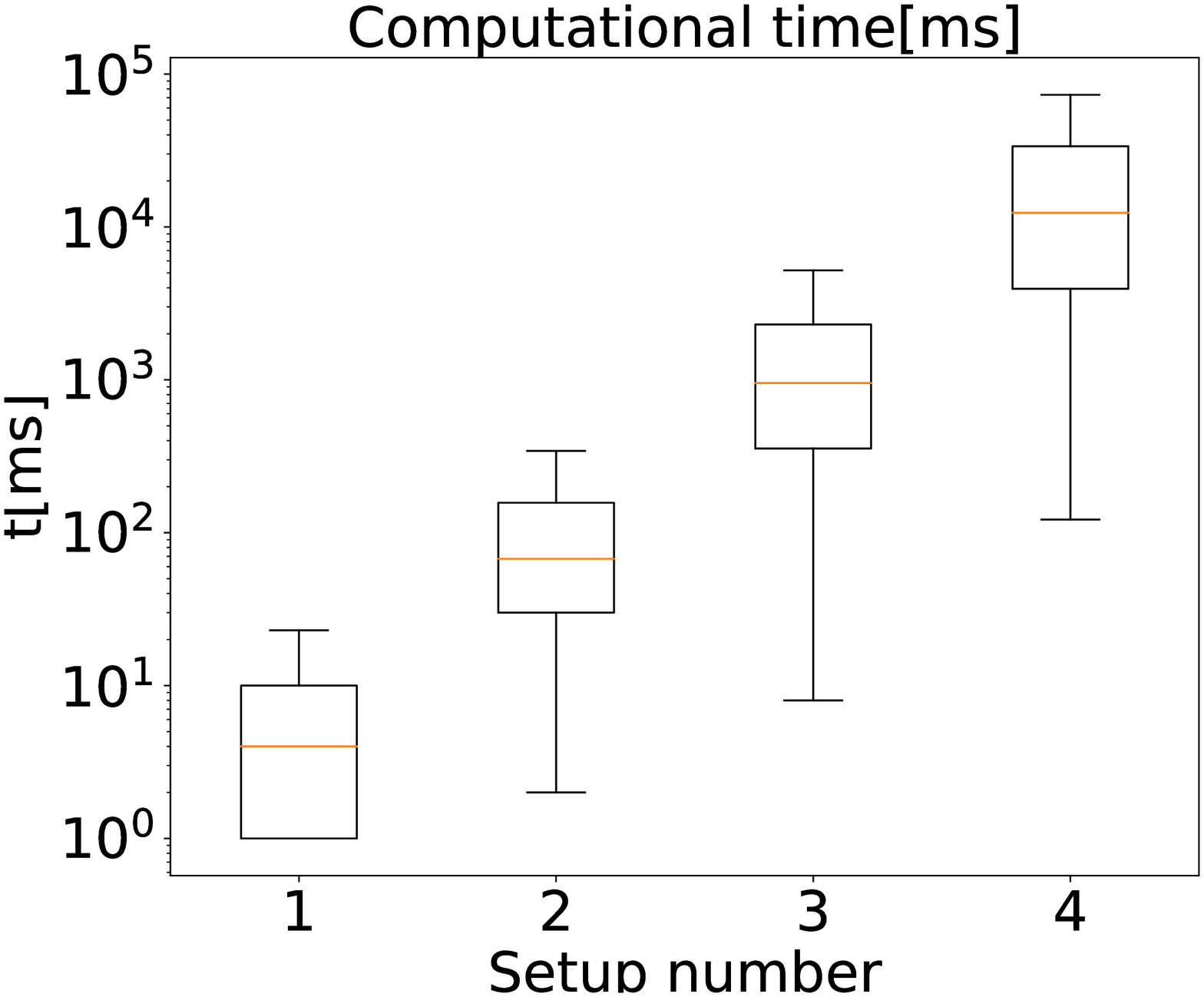} 
\includegraphics[width=5cm,scale=1]{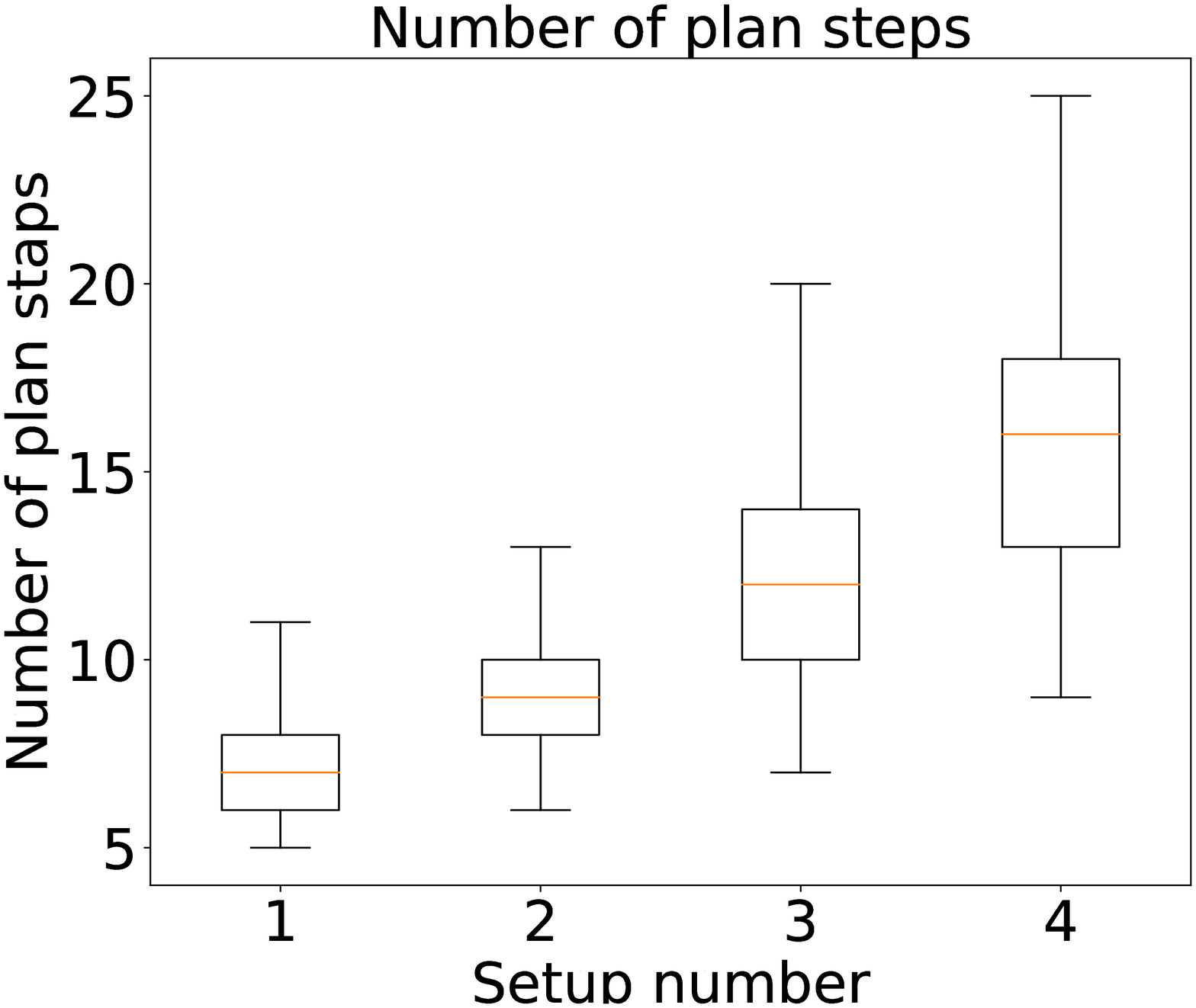} \\
\end{tabular}
\caption{Results of the proposed approach on assignments which CARP is unable to solve.}
\label{fig:second_experiment}
\end{figure*}

The second set of experiments was carried out on the second set of maps with the aim to illustrate the behavior of the proposed algorithm on assignments that CARP algorithm can not solve.
The total of 400 different combinations of a map and assignment were generated: 100 each for 10, 20, 30 and 40 agents.
The results of this experiment can be seen in Fig.~\ref{fig:second_experiment}.
The setup numbers 1 to 4 correspond to the number of agents 10 to 40 respectively. 

For up to 30 agents the success rate is 100\% while it is decreased to 95\% for 40 robots.
Regarding the computational time results, the algorithm takes approximately 1 second to calculate the paths for each agent in assignments that are impossible to solve for CARP algorithm for up to 30 agents even with relatively complicated assignments. 

\section{\uppercase{Conclusion}}
\label{sec:conclusion}

In this paper, we presented a novel algorithm for coordination of a fleet of robots on a graph. 
The algorithm is based on a discrete version of RRT for multiple robots (MRdRRT), but it significantly improves particular steps of this algorithm which allows it to solve problems assuming tens of robots in few seconds. 
This is in contrast to the original MRdRRT which can solve problems up to ten robots in tens of seconds.
The experimental comparison moreover shows that the proposed approach can solve problems unsolvable for CARP which is one of the best practical algorithms nowadays.  
Finally, our approach is comparable to CARP in computational time and quality of the generated solutions for problems with up to 100 robots which are solvable by CARP. The main drawback is that the computational complexity with respect to the number of robots is still exponential.

The ideas of RRT* showed promise in bringing the obtained solution closer to optimum, but for the cost of increased execution time. 
For this reason, the future work should focus on improvement of the proposed algorithm in terms of reducing the number of required iterations
to find the first solution. This could be done for example by reducing the dimensionality of the problem by planning smaller groups of agents in batches
and then considering them as obstacles moving in time for the subsequent groups.

\section*{\uppercase{Acknowledgements}}

\noindent This work has been supported by the European Union's Horizon 2020 research and innovation programme under grant agreement No 688117, by the Technology Agency of the Czech Republic under the project no.~TE01020197 \enquote{Centre for Applied Cybernetics}, the project Rob4Ind4.0 CZ.02.1.01/0.0/0.0/15\_003/0000470 and the European Regional Development Fund. 
The work of Jakub Hv\v{e}zda was also supported by the Grant Agency of the Czech Technical University in Prague, grant No.~SGS18/206/OHK3/3T/37.

\bibliographystyle{apalike}
{\small
\bibliography{main}}

\end{document}